\documentclass{article}
\usepackage{spconf,amsmath,graphicx}

\usepackage{graphicx}
\usepackage{booktabs}
\usepackage{amsmath,amsfonts,amssymb}
\usepackage{multirow}
\usepackage{algorithm}
\usepackage{algorithmic}
\usepackage{subfigure}
\usepackage{amsthm}
\usepackage{url}            
\usepackage{amsfonts}       
\usepackage{amsmath}
\newcommand{\xvec}{{\bf x}}

\title{Mixup Regularized Adversarial Networks for Multi-Domain Text Classification}
\name{Yuan Wu$^{\star}$ \qquad Diana Inkpen$^{\dagger}$ \qquad Ahmed El-Roby$^{\star}$}
\address{$^{\star}$ Carleton University, Ottawa, Canada \\
      $^{\dagger}$ University of Ottawa, Ottawa, Canada}
\begin{document}
%
\maketitle
\begin{abstract}
Using the shared-private paradigm and adversarial training has significantly improved the performances of multi-domain text classification (MDTC) models. However, there are two issues for the existing methods. First, instances from the multiple domains are not sufficient for domain-invariant feature extraction. Second, aligning on the marginal distributions may lead to fatal mismatching. In this paper, we propose a mixup regularized adversarial network (MRAN) to address these two issues. More specifically, the domain and category mixup regularizations are introduced to enrich the intrinsic features in the shared latent space and enforce consistent predictions in-between training instances such that the learned features can be more domain-invariant and discriminative. We conduct experiments on two benchmarks: The Amazon review dataset and the FDU-MTL dataset. Our approach on these two datasets yields average accuracies of 87.64\% and 89.0\% respectively, outperforming all relevant baselines.

\end{abstract}
%

\section{Introduction}
Text classification is a fundamental task in natural language processing (NLP) and has been successfully applied in a wide variety of applications, such as spam detection \cite{ngai2011application}, fake news detection \cite{shu2017fake} and sentiment analysis \cite{yang2019multi}. However, the impressive achievements of text classification are based on large amounts of annotated training data. In real-world scenarios, the availability of annotated data can largely vary among domains. For certain popular domains like movie, book, and electronic product reviews, manually-annotated datasets have been made available. While for domains such as medical equipment reviews, there only exist limited amounts of annotated data. Therefore, it is of great importance to investigate how to improve the classification accuracy on the interested domain by leveraging annotated data from related domains.

Multi-domain text classification (MDTC) is proposed to address the above problem \cite{wu2020dual2}. Currently, most state-of-the-art MDTC methods adopt the shared-private paradigm \cite{bousmalis2016domain} and adversarial training \cite{goodfellow2014generative} to improve the system performance. The shared-private paradigm constructs two types of feature spaces: the shared latent space is used to capture the common features across different domains, while the domain-specific one aims to extract domain-specific features. The adversarial training was first proposed for image generation \cite{goodfellow2014generative}, then it was extended to align different domains to extract domain-invariant features for domain adaptation \cite{ganin2016domain}. The idea behind integrating adversarial training with MDTC is to perform domain alignments and guarantee the optimum separations among the shared latent space and multiple domain-specific feature spaces, preventing domain-specific features from emerging into the shared latent space. However, these approaches still face two issues: First, since we always use mini-batch stochastic gradient descent for optimization, if the batch size is small, instances drawn from multiple domains may not be sufficient to guarantee strong domain-invariance to the shared latent space. Second, aligning on the marginal distributions may lead to fatal mismatching, resulting in weak discriminability of the learned features \cite{kumar2018co}.

%

\begin{figure*}
    \centering
    \includegraphics[width=1.1\columnwidth]{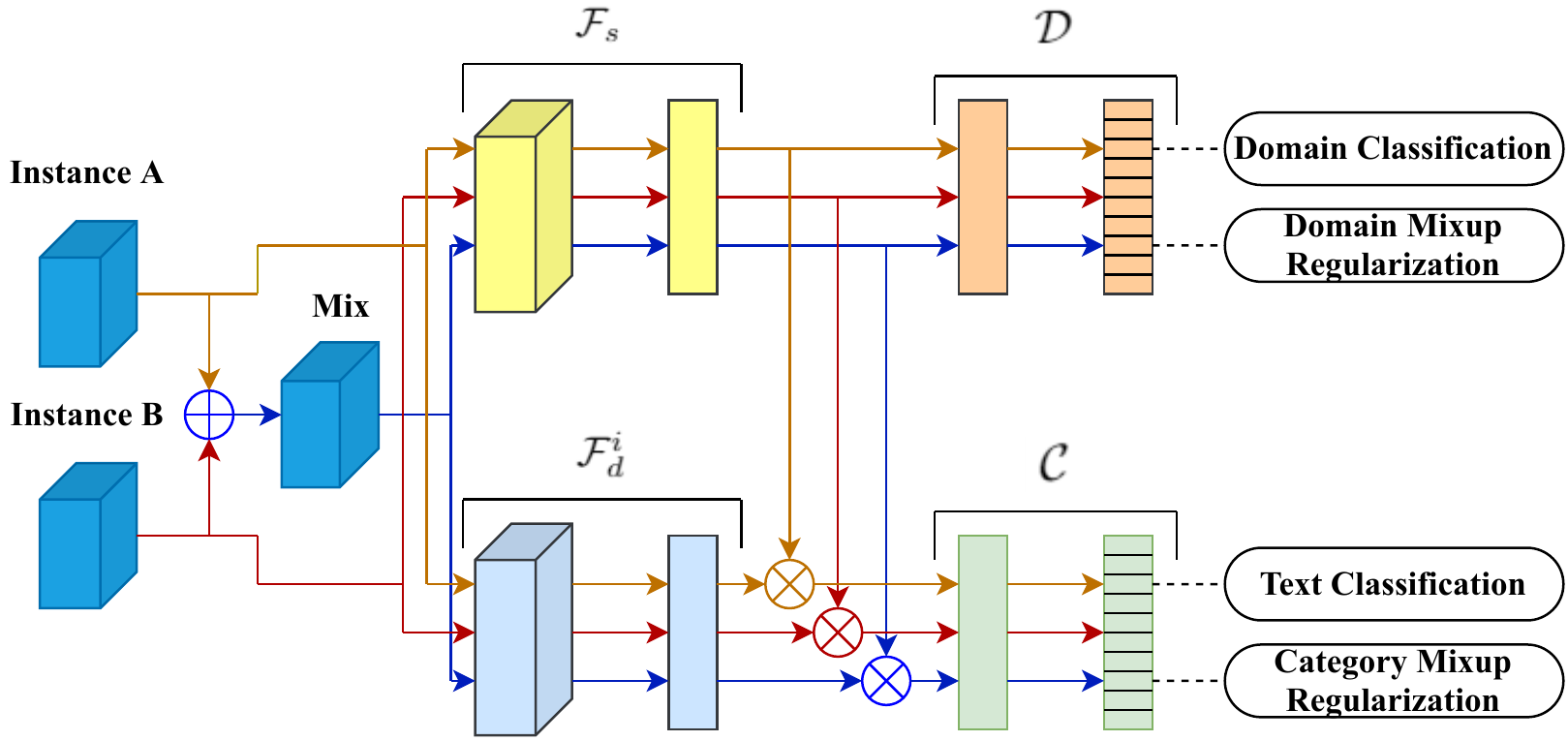}
    \caption{Architecture of the mixup regularized adversarial network model. 
	A shared feature extractor $\mathcal{F}_s$ learns to capture domain-invariant features;  
	each domain-specific feature extractor $\mathcal{F}^i_{d}$
	learns to capture domain-specific features;  
	a multinomial domain discriminator $\mathcal{D}$ is used to discriminate shared features across different domains;  
	and a classifier $\mathcal{C}$ is used to conduct text classification.
	}
    \label{Fig2}
\end{figure*}


In this paper, we propose mixup regularized adversarial networks (MRANs), which conduct mixup-based interpolation regularizations to tackle the aforementioned two issues for MDTC. Mixup \cite{zhang2018mixup} is a form of vicinal risk minimization, which trains models on convex combinations of training instances and labels to combat memorization of corrupted labels and instability of adversarial training. Motivated by mixup, this paper makes the key contribution by introducing two types of mixup-based interpolation regularizations: domain mixup regularization and category mixup regularization. More specifically, the domain mixup regularization can expand the searching range in the shared latent space, excavating more intrinsic feature patterns such that a more continuous shared latent space can be achieved. The category mixup regularization can enforce consistent predictions on the linear interpolations for both labeled and unlabeled instances, imposing stronger discriminability to the latent spaces. These two regularizations enable the proposed MRAN to avoid mismatching and generalize well on each domain. Experiments are carried out on two MDTC benchmarks: the Amazon Review dataset and the FDU-MTL dataset. The proposed method can outperform the state-of-the-art methods on both datasets.


\section{The Approach}

In this paper, the MDTC tasks are presented in a real-world setting in which texts come from various domains, each with a varying amount of annotated instances. Specifically, assume there exist $M$ domains $\{D_i\}_{i=1}^{M}$, each domain $D_i$ consists of two parts: a limited amount of annotated instances $\mathbb{L}_i=\{(\xvec_j,y_j)\}_{j=1}^{l_i}$; and a set of unlabeled instances $\mathbb{U}_i=\{\xvec_j\}_{j=1}^{u_i}$. The task is to train a model that can map an instance $\xvec$ to its corresponding label $y$. The ultimate goal is to improve the overall performance of the system, measured in this paper as the average classification accuracy across the $M$ domains.


\subsection{Adversarial Training}

As illustrated in Figure \ref{Fig2}, our proposed MRAN can be decomposed into four components: a shared feature extractor $\mathcal{F}_s$, a set of domain-specific feature extractors $\{\mathcal{F}_d^i\}_{i=1}^M$, a multinomial domain discriminator $\mathcal{D}$, and a classifier $\mathcal{C}$. In MRAN, the input data should be processed by two data flows: one is to extract the generalizable features which can enable knowledge sharing across all domains, while the other one is to capture domain-specific features that can maximally represent useful information for their own domain. These two flows complement each other. The domain discriminator $\mathcal{D}$ takes a domain-invariant feature vector as its input and outputs the likelihood of the given instance coming from each domain. For $M$ domains, $\mathcal{D}$ will be a $M$-class classifier and we use $\mathcal{D}_i(\mathcal{F}_s(\xvec))$ to denote the predicted probability of instance $\xvec$ coming from the $i$-th domain. A two-player mini-max game is played between $\mathcal{D}$ and $\mathcal{F}_s$, where $\mathcal{D}$ tries to distinguish features from different domains, while $\mathcal{F}_s$'s goal is to maximally confuse $\mathcal{D}$. We use the negative log-likelihood (NLL) loss to encode the multinomial adversarial loss function as follows:

\begin{align}
	\mathcal{L}_{Adv}=&
	-\sum_{i=1}^M \mathbb{E}_{\xvec_i\sim\mathbb{L}_i\cup\mathbb{U}_i} 
	{\log[ \mathcal{D}_i(\mathcal{F}_s(\xvec_i))]}
	\label{eq:advD}
\end{align}

After this training, the shared feature extractor $\mathcal{F}_s$ and the discriminator $\mathcal{D}$ can reach an equilibrium at which both $\mathcal{D}$ and $\mathcal{F}_s$ cannot improve and the learned features are domain-invariant.

The NLL is also employed to encode the classification loss. The classifier $\mathcal{C}$ takes the concatenation of a shared feature and a domain-specific feature as its input vector and outputs the label probability of the given instance. It utilizes the multiple layer perceptrons (MLPs) with a softmax layer on top. The classifier $\mathcal{C}$ is trained together with the feature extractors by minimizing the NLL on the annotated instances:

\begin{align}
    \mathcal{L}_c = 
	-{\sum_{i=1}^M \mathbb{E}_{(\xvec_i, y_i)\sim{\mathbb{L}_i}}} 
	\log[\mathcal{C}_{y_i}(\mathcal{F}^i(\xvec_i)]
\end{align}

\noindent where $\mathcal{F}^i(\xvec)$ is the concatenation of $\mathcal{F}_s(\xvec)$ and $\mathcal{F}_d^i(\xvec)$, $\mathcal{C}_{y}(\mathcal{F}^i(\xvec))$ denotes the probability of instance $\xvec$ belonging to the category $y$.

\subsection{Mixup-based Regularization}

The limited amounts of annotated data in each domain may lead to overfitting and aligning on the marginal distributions may lead to mismatching. In order to alleviate these harmful effects, we propose two mixup-based interpolation regularizations. Mixup is an effective and data-agnostic data augmentation technique, it can address issues such as memorization of corrupted labels and instability of adversarial training \cite{zhang2018mixup}. Mixup constructs virtual instances as the linear interpolation of two random instances $(\xvec^k, \xvec^s)$ from the training set, and implements the same operation on their corresponding labels $(y^k, y^s)$ to generate pseudo-labels.

\begin{align}
    \widetilde{\xvec} = & 
    \mathcal{M}_\lambda(\xvec^k,\xvec^s)=\lambda \xvec^k+(1-\lambda)\xvec^s
    \\
    \widetilde{y} = &
    \mathcal{M}_\lambda(y^k,y^s)=\lambda y^k+(1-\lambda)y^s
\end{align}

\noindent where $\lambda$ is sampled from a beta distribution $Beta(\alpha,\alpha)$ for $\alpha\in(0,\infty)$. Mixup has been proven effective in both the input space and the latent space to improve the system performance for various tasks \cite{zhang2018mixup,verma2019manifold}. In MRAN, we conduct category and domain mixup regularizations in the input space.

For the category mixup regularization, we first apply mixup on the annotated instances in each domain to enforce prediction consistency:

\begin{equation}
    \begin{aligned}
    \mathcal{L}_a^M=&
    -\sum_{i=1}^M\mathbb{E}_{(\xvec_i^k,y_i^k),(\xvec_i^s,y_i^s)\sim\mathbb{L}_i}\log[\mathcal{C}_{\widetilde{y}_i}(\mathcal{F}^i(\widetilde{\xvec}_i)]
\end{aligned}
\end{equation}

Then, for the unlabeled instances, since we have no direct access to the label information. We use pseudo-labels in lieu of real labels and mixup is applied on the instances and the generated pseudo-labels. In order to guarantee the linear behavior of the model in-between training instances, we introduce a penalty term which punishes the difference between the prediction of interpolation of two instances and the interpolation of predictions of these instances:

\begin{equation}
    \begin{aligned}
    \mathcal{L}_u^M=
    \sum_{i=1}^M\mathbb{E}_{(\xvec^k_i,\xvec^s_i)\sim\mathbb{U}_i}Dis(p(\widetilde{x}_i),\mathcal{M}_\lambda(p(\xvec^k_i),p(\xvec^s_i)))
    \end{aligned}
\end{equation}

\noindent where $p(\xvec)=\log[C(\mathcal{F}^i(\xvec))]$, $C(\cdot)$ denotes the softmax prediction of the classifier $\mathcal{C}$, $Dis(\cdot,\cdot)$ is the penalty term. In our experiments we use the $\ell_1$-norm function as the discrepancy metric of $Dis(\cdot,\cdot)$. The category mixup regularization can enforce prediction consistency in-between training instances to enhance the discriminability of the learned features.

For the domain mixup regularization, we conduct mixup interpolations per domain. Since the domain label of the instances sampled from one domain should be identical, it is not necessary to generate their pseudo domain labels. The domain mixup regularization is defined as follows:

\begin{align}
    \mathcal{L}_{Adv}^M=&
    -\sum_{i=1}^M\mathbb{E}_{(\xvec^k_i,\xvec^s_i)\sim\mathbb{L}_i\cup\mathbb{U}_i} \log[\mathcal{D}_i(\mathcal{F}_s(\widetilde{\xvec}_i))]
\end{align}

\noindent where the mixup ratio $\lambda$ is the same as the one used in the category mixup regularization. The domain mixup regularization enriches the intrinsic feature patterns in the shared latent space such that a more continuous shared latent space can be achieved. Finally, the MRAN method can be formulated as: 

\begin{equation}
    \begin{aligned}
        \min_{\mathcal{F}_s,\{\mathcal{F}_d^i\}_{i=1}^M, C}\max_{\mathcal{D}} \quad
	&\mathcal{L}_c+\lambda_d\mathcal{L}_{Adv}+ \\
	&\lambda_a\mathcal{L}_a^M+\lambda_u\mathcal{L}_u^M+\lambda_m\mathcal{L}_{Adv}^M
    \end{aligned}
\end{equation}

\noindent where $\lambda_d$, $\lambda_a$, $\lambda_u$ and $\lambda_m$ are hyperparameters for balancing different losses. 


\section{Experiments}

\subsection{Dataset}


In our experiments, two benchmarks are used to empirically demonstrate the effectiveness of our MRAN: the Amazon review dataset \cite{blitzer2007biographies} and the FDU-MTL dataset \cite{liu2017adversarial}. There are four domains in the Amazon review dataset: books, DVDs, electronics, and kitchen. Each domain has 2,000 samples: 1,000 positive ones and 1,000 negative ones. This dataset was pre-processed into a bag of features (unigrams and bigrams), losing all word order information. We take the 5,000 most frequent features and represent each review as a 5,000-dimensional vector. For the FDU-MTL dataset, it consists of 16 domains: 14 Amazon review domains (books, electronics, DVDs, kitchen, apparel, camera, health, music, toys, video, baby, magazine, software, and sport) and two movie reviews (IMDB and MR). The reviews in FDU-MTL are raw text data being tokenized by the Stanford tokenizer. Since we cannot directly mix the instances if they are raw text data, we conduct mixing on the embeddings obtained by processing each instance via word2vec \cite{mikolov2013efficient}. Each domain has a development set of 200 samples and a test set of 400 samples. The numbers of labeled and unlabeled samples vary across domains but are roughly 1,400 and 2,000, respectively.

\subsection{Comparison Methods}

To comprehensively evaluate the efficacy of our method, we list several baselines. The collaborative multi-domain sentiment classification (CMSC) combines a classifier shared across all domains with a set of classifiers, one per domain, to make the final prediction \cite{wu2015collaborative}. The CMSC models can be trained on three different loss functions: the least square loss (CMSC-LS), the hinge loss (CMSC-SVM), and the log loss (CMSC-Log). The adversarial multi-task learning for text classification (ASP-MTL) uses adversarial learning and orthogonality constraints to guide feature extraction \cite{liu2017adversarial}. The multinomial adversarial networks (MANs) adopt the adversarial learning and shared-private paradigm, exploiting two forms of loss functions to train the multinomial domain discriminator: the least square loss (MAN-L2) and the negative log-likelihood loss (MAN-NLL) \cite{chen2018multinomial}. The multi-task learning with bidirectional language (MT-BL) method utilizes language modeling as an auxiliary task for domain-specific feature extraction, aiming to enhance its ability to capture domain-specific features \cite{yang2019multi}. All the comparison methods use the standard partitions of the datasets. Thus, we cite the results from \cite{chen2018multinomial,yang2019multi} for fair comparisons.

\subsection{Implementation Details}

In our experiments, we set hyperparameters $\alpha=0.2$, $\lambda_d=1$, $\lambda_a=0.001$, $\lambda_u=0.1$ and $\lambda_m=0.00001$, use the Adam optimizer \cite{kingma2014adam}, with the learning rate 0.0001, for training and report results based on three random experiments. The size of the domain-invariant feature vector is 128 while the size of the domain-specific feature vector is 64. The dropout rate for each component is 0.4. $\mathcal{C}$ and $\mathcal{D}$ are MLPs with one hidden layer of the same size as their input ($128+64$ for $\mathcal{C}$ and $128$ for $\mathcal{D}$). ReLU is used as the activation function. The batch size is 8. For the experiments on the Amazon review dataset, MLP feature extractors are used (the input size is 5,000). Each extractor has two hidden layers, with size 1,000 and 500, respectively. For the experiments on the FDU-MTL dataset, we adopt a CNN feature extractor with a single convolutional layer. It uses different kernel sizes (3, 4, 5), and the number of kernels is 200. The input of the convolutional layer is the 100-dimensional word embedding, obtained by using word2vec \cite{mikolov2013efficient}, for each word in the input sequence.

\subsection{Results}


\begin{table}[t]
\caption{\label{font-table} MDTC results on the Amazon review dataset.}\smallskip
\label{table_ref1}
\centering
\resizebox{1.0\columnwidth}{!}{
\smallskip\begin{tabular}{ l|  c c c c c c c}
\hline
	Domain & CMSC-LS & CMSC-SVM & CMSC-Log & MAN-L2 & MAN-NLL & MRAN(Proposed)\\
\hline
Books &  82.10 & 82.26 & 81.81 & 82.46 & 82.98 & $\mathbf{84.60\pm0.18}$ \\
DVD &  82.40 & 83.48 & 83.73 & 83.98 & 84.03 & $\mathbf{85.60\pm0.16}$ \\
Electr.  & 86.12 & 86.76 & 86.67 & 87.22 & 87.06 & $\mathbf{89.10\pm0.05}$ \\
Kit.  &  87.56 & 88.20 & 88.23 & 88.53 & 88.57 & $\mathbf{91.25\pm0.19}$\\
\hline
AVG  &  84.55 & 85.18 & 85.11 & 85.55 & 85.66 & $\mathbf{87.64\pm0.08}$\\
\hline
\end{tabular}}
\end{table}
\begin{table}[t]
\caption{\label{font-table} MDTC results on the FDU-MTL dataset.}\smallskip
\label{table_ref2}
\centering
\resizebox{1.0\columnwidth}{!}{
\begin{tabular}{ l| c c c c c}
\hline
	Domain & ASP-MTL & MAN-L2 & MAN-NLL & MT-BL & MRAN(Proposed)\\
\hline
books & 84.0 & 87.6 & 86.8 & $\mathbf{89.0}$ & 87.0$\pm$0.2 \\
electronics & 86.8 & 87.4 & 88.8 & $\mathbf{90.2}$ & 89.0$\pm$0.7 \\
dvd & 85.5 & 88.1 & 88.6 & 88.0 & $\mathbf{89.0\pm0.4}$ \\
kitchen & 86.2 & 89.8 & 89.9 & 90.5 & $\mathbf{93.0\pm0.2}$\\
apparel & 87.0 & 87.6 & 87.6 & 87.2 & $\mathbf{91.5\pm0.3}$\\
camera & 89.2 & 91.4 & 90.7 & 89.5 & $\mathbf{93.0\pm0.4}$\\
health & 88.2 & 89.8 & 89.4 & $\mathbf{92.5}$ & 90.0$\pm$0.3 \\
music & 82.5 & 85.9 & 85.5 & 86.0 & $\mathbf{86.5\pm0.1}$ \\
toys & 88.0 & 90.0 & 90.4 & $\mathbf{92.0}$ & 86.0$\pm$0.3 \\
video & 84.5 & 89.5 & $\mathbf{89.6}$ & 88.0 & 88.5$\pm$0.4\\
baby & 88.2 & 90.0 & $\mathbf{90.2}$ & 88.7 & 90.0$\pm$0.5 \\
magazine & 92.2 & 92.5 & 92.9 & 92.5 & $\mathbf{93.5\pm0.3}$ \\ 
software & 87.2 & 90.4 & 90.9 & $\mathbf{91.7}$ & 89.5$\pm$0.4 \\
sports & 85.7 & 89.0 & 89.0 & 89.5 & $\mathbf{90.5\pm0.2}$ \\
IMDb & 85.5 & 86.6 & 87.0 & 88.0 & $\mathbf{89.0\pm0.1}$\\
MR & 76.7 & 76.1 & 76.7 & 75.7 & $\mathbf{78.5\pm0.5}$\\
\hline
AVG & 86.1 & 88.2 & 88.4  & 88.6  & $\mathbf{89.0\pm0.1}$ \\
\hline
\end{tabular} 
}
\end{table}

The experiments on the Amazon review dataset follow the setting of \cite{wu2015collaborative}. The 5-fold cross-validation is conducted. We divide the data into five folds per domain: three of the five folds are used for the training set, one is the validation set, and the remaining one is treated as the test set. The experimental results on the Amazon review dataset are shown in Table \ref{table_ref1},  we can see that our model achieves the highest average accuracy of $87.64\%$, surpassing MAN-NLL by a significant margin of $1.98\%$. Moreover, our model performs best on each individual domain comparing with other baselines.

From the experimental results on the FDU-MTL dataset, reported in Table \ref{table_ref2}, we can see that, in terms of average accuracy, the performance of our proposed MRAN is superior to the other methods. Moreover, our model performs best on 9 of the 16 domains. The best classification accuracies on books, electronics, health, toys, and software are obtained by MT-BL. However, MT-BL adds language modeling as an auxiliary task for domain-specific feature extraction by using an extra bidirectional long short-term memory network, which makes the model architecture more complicated. The experimental results on the FDU-MTL dataset again illustrate the effectiveness of our model for MDTC. 


\subsection{Ablation Study}

\begin{table}
\caption{\label{font-table} Ablation study analysis.}\smallskip
\label{table_ref4}
\centering
\resizebox{1.0\columnwidth}{!}{
\begin{tabular}{ l| c c c c c }
\hline
Method & Books & DVD & Electr. & Kit. & AVG\\
\hline
MRAN (full)& 84.60 & 85.60 & 89.10 & 91.25 & 87.64 \\
MRAN w/o DM & 84.05 & 85.35 & 87.50 & 90.50 & 86.85 \\
MRAN w/o CM & 83.85 & 84.90 & 88.00 & 89.20 & 86.49 \\
MRAN w/o LCM & 83.60 & 85.00 & 87.20 & 90.25 & 86.51 \\
MRAN w/o UCM & 84.10 & 85.55 & 86.70 & 89.80 & 86.54 \\
\hline
\end{tabular}}
\end{table}

In order to verify the contributions of domain mixup (DM) regularization and category mixup (CM) regularization, we conduct an extensive ablation study on the Amazon review dataset. In addition, for the category mixup regularization, we both evaluate the contributions of the category mixup regularization on labeled data (LCM) and the category mixup regularization on unlabeled data (UCM). From Table \ref{table_ref4}, it can be noted that all four variants obtain inferior results, and the full MRAN produces the best results. The model without CM performs the worst, which shows that CM contributes more to system performance than the DM part. Moreover, both LCM and UCM, which encourage prediction consistency, contribute to the improved performance of our model. 


\section{Conclusion}

In this paper, we propose mixup regularized adversarial networks (MRANs) for multi-domain text classification. The proposed MRAN introduces the category mixup regularization to enforce consistent predictions in-between training instances, imposing stronger discriminability to the latent spaces, and the domain mixup regularization to explore more internal features in the shared latent space, leading to a more continuous shared latent space. These two mixup-based interpolation regularizations can enhance and complement each other. The experiments on two MDTC benchmarks show that our MRAN can effectively improve system performance.


\bibliographystyle{IEEEbib}
\bibliography{strings}

\end{document}